\documentclass[9pt,sigconf,nonacm]{acmart}

\usepackage[frozencache,cachedir=.]{minted}
\usepackage{hyperref}
\usepackage{threeparttable}
\usepackage{array}
\usepackage{multirow}
\usepackage{svg}
\usepackage{fontawesome5}

\newcommand{\figref}[1]{Figure~\ref{#1}}
\newcommand{\secref}[1]{Section~\ref{#1}}
\newcommand{\tabref}[1]{Table~\ref{#1}}

\newcolumntype{P}[1]{>{\centering\arraybackslash}m{#1}}

\AtBeginDocument{%
  \providecommand\BibTeX{{%
    \normalfont B\kern-0.5em{\scshape i\kern-0.25em b}\kern-0.8em\TeX}}}

\everypar{\looseness=-1}
\newcounter{BalanceAtReference}
\newcounter{ReferenceIndexForBalancing}

\makeatletter

\global\@ACM@balancefalse

\def\@balancelastpageonce{%
  \ifnum\value{ReferenceIndexForBalancing}=\value{BalanceAtReference}
    \newpage
  \else
    \relax
  \fi
  \stepcounter{ReferenceIndexForBalancing}
}
\pretocmd{\bibitem}{\@balancelastpageonce}
  {} % on success
  {\@latex@error{Patching \bibitem failed}{\@ehd}}

\makeatother

\copyrightyear{2023}
\acmYear{2023}
\setcopyright{acmlicensed}\acmConference[SAFE '23]{Proceedings of the 2023 Explainable and Safety Bounded, Fidelitous, Machine Learning for Networking}{December 8, 2023}{Paris, France}
\acmBooktitle{Proceedings of the 2023 Explainable and Safety Bounded, Fidelitous, Machine Learning for Networking (SAFE '23), December 8, 2023, Paris, France}
\acmPrice{15.00}
\acmDOI{10.1145/3630050.3630176}
\acmISBN{979-8-4007-0449-9/23/12}
\begin{document}

%%
%% The "title" command has an optional parameter,
%% allowing the author to define a "short title" to be used in page headers.

%\title{DataZoo: Streamlining development of traffic classifiers}
% \title{DataZoo: Simplified evaluation of traffic classifiers}
\title{DataZoo: Streamlining Traffic Classification Experiments}

%%
%% The "author" command and its associated commands are used to define
%% the authors and their affiliations.
%% Of note is the shared affiliation of the first two authors, and the
%% "authornote" and "authornotemark" commands
%% used to denote shared contribution to the research.

\author{Jan Luxemburk}
\email{luxemburk@cesnet.cz}
\affiliation{%
 \institution{\textit{FIT CTU \& CESNET}}
 \city{Prague}
 \country{Czech Republic}
}

\author{Karel Hynek}
\email{hynekkar@cesnet.cz}
\affiliation{%
 \institution{\textit{FIT CTU \& CESNET}}
 \city{Prague}
 \country{Czech Republic}
}

\renewcommand{\shortauthors}{Jan Luxemburk \& Karel Hynek}

\begin{abstract}
The machine learning communities, such as those around computer vision or natural language processing, have developed numerous supportive tools and benchmark datasets to accelerate the development. In contrast, the network traffic classification field lacks standard benchmark datasets for most tasks, and the available supportive software is rather limited in scope. This paper aims to address the gap and introduces DataZoo, a toolset designed to streamline dataset management in network traffic classification and to reduce the space for potential mistakes in the evaluation setup. DataZoo provides a standardized API for accessing three extensive datasets---CESNET-QUIC22, CESNET-TLS22, and CESNET-TLS-Year22. Moreover, it includes methods for feature scaling and realistic dataset partitioning, taking into consideration temporal and service-related factors. The DataZoo toolset simplifies the creation of realistic evaluation scenarios, making it easier to cross-compare classification methods and reproduce results.
\end{abstract}

%%
%% The code below is generated by the tool at http://dl.acm.org/ccs.cfm.
%% Please copy and paste the code instead of the example below.
%%

\begin{CCSXML}
<ccs2012>
<concept>
<concept_id>10002978.10003014</concept_id>
<concept_desc>Security and privacy~Network security</concept_desc>
<concept_significance>300</concept_significance>
</concept>
<concept>
<concept_id>10010147.10010257</concept_id>
<concept_desc>Computing methodologies~Machine learning</concept_desc>
<concept_significance>300</concept_significance>
</concept>
<concept>
<concept_id>10003033.10003079.10011704</concept_id>
<concept_desc>Networks~Network measurement</concept_desc>
<concept_significance>300</concept_significance>
</concept>
</ccs2012>
\end{CCSXML}

\ccsdesc[300]{Security and privacy~Network security}
\ccsdesc[300]{Computing methodologies~Machine learning}
\ccsdesc[300]{Networks~Network measurement}

%%
%% Keywords. The author(s) should pick words that accurately describe
%% the work being presented. Separate the keywords with commas.
\keywords{Traffic classification, Open datasets, Application identification, Machine learning, Open-world evaluation, Toolset, Encrypted traffic, TLS, QUIC}

% \received{20 February 2007}
% \received[revised]{12 March 2009}
% \received[accepted]{5 June 2009}

%%
%% This command processes the author and affiliation and title
%% information and builds the first part of the formatted document.
\maketitle

\vspace{0.5cm}
\hspace{-0.5cm}
\includegraphics[width=0.75\columnwidth]{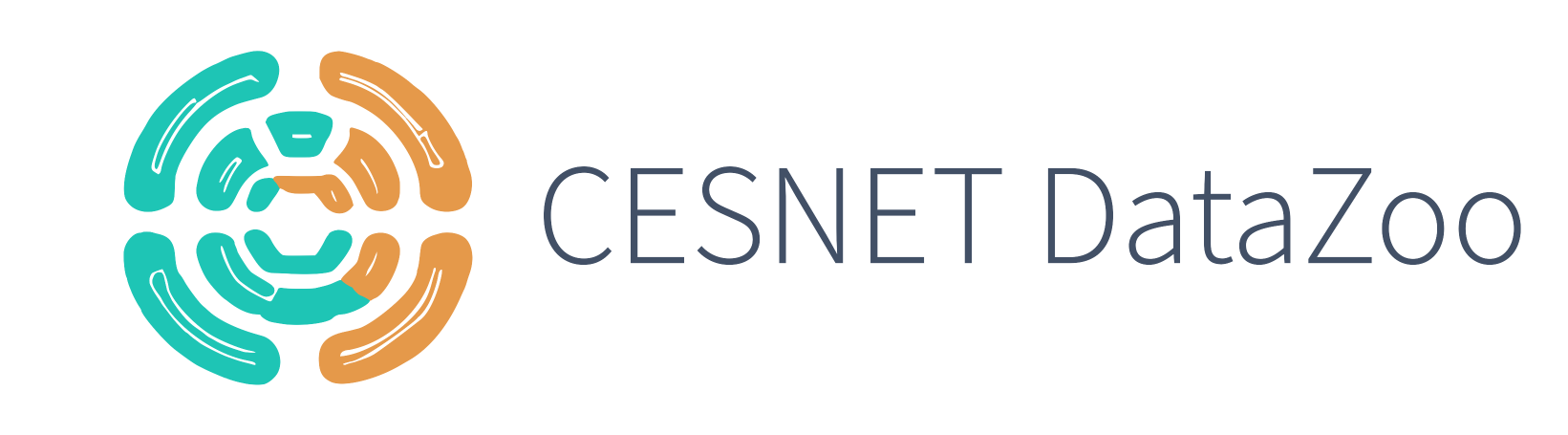}

\noindent
\faGithubSquare\ GitHub  -- \url{https://github.com/CESNET/cesnet-datazoo} \\
\faSearch\ Documentation -- \url{https://cesnet.github.io/cesnet-datazoo} \\
\faCube\ Install with \textcolor{red}{\texttt{pip install cesnet-datazoo}} \\
\textit{Contact us with issues, we appreciate feedback.}
%\textit{Please contact us if you encounter any issues; we appreciate feedback.}
\vspace{1cm}

\begin{table*}[ht!]
    \centering
    \caption{The detailed information about datasets included in the initial release of the DataZoo toolset.}
    \label{tab:datasets}
    \begin{tabular}{|l|l|l|l|}
    \hline
        \textbf{Name} &\textbf{CESNET-TLS22} & \textbf{CESNET-QUIC22} & \textbf{CESNET-TLS-Year22} \\ \hline
        \textbf{Protocol} & TLS & QUIC & TLS \\ \hline
        \textbf{Published} in & 2022 & 2023 & To be published in Q4 2023 \\ \hline
        \textbf{Collection duration} & 2 weeks & 4 weeks & 1 year \\ \hline
        \textbf{Collection period} & 4.10.2021 - 17.10.2021 & 31.10.2022 - 27.11.2022 & 1.1.2022 - 31.12.2022 \\ \hline
        \textbf{Class labels} & 191 & 102 & 182 \\ \hline
        \textbf{Available samples} & 141M & 153M & 507M \\ \hline
    \end{tabular}
\end{table*}

\section{Introduction and Related Works}

Machine learning (ML) is a core technology for the classification of network traffic. The ML-based models utilize sequences of packet sizes, times, and other connection metadata statistics to perform classification and threat detection even in encrypted traffic. Nevertheless, further adoption of ML and its development is being slowed down because of the lack of representative datasets and supportive tools~\cite{Wang2022}. As a result, traffic classification experiments require more effort, are less reproducible, and are prone to unintentional errors.

Mistakes in the evaluation setup, which can increase the measured ML model performance, are still prevalent in the network traffic classification field~\cite{Arp2022}. For example, evaluation should be time-consistent in the sense that test data have to come after train data, which is often not the case (for example, when a dataset is shuffled in k-fold cross-validation). Not respecting the time order of train and test sets can lead to results tens-percent higher than the actual time-consistent results, as was demonstrated for Android malware classification~\cite{pendlebury2019}. Another common problem was described in~\cite{Yang2021}---when the number of classes is too small, the classification problem is often trivial to solve, and the reported performance might be biased.

Achieving realistic evaluation, which includes all the nuances of real deployment, is a challenging task due to the constant changes in the network environment---devices are connected and disconnected, traffic characteristics differ between rush and off hours, network service updates can change the traffic shape, or sudden appearances of novel network services. All of these factors contribute to data drift, a phenomenon causing a reduced ML-based classifier performance when deployed to a real network.

In this paper, we introduce DataZoo---a Python toolset for streamlining the work with large network traffic datasets that reduces the space for potential mistakes in the evaluation setup. The toolset provides a common API for obtaining three extensive datasets (CESNET-QUIC22~\cite{Luxemburk2023QUICDataset}, CESNET-TLS22~\cite{Luxemburk2023}, and CESNET-TLS-Year22), and more will come in the future. The datasets are available in multiple sizes to let users start experimenting at a smaller scale or perform evaluation on full datasets with up to 500 million samples. 

The motivation behind DataZoo stems from the lack of standard benchmarks and supportive tools targeting network traffic classification~\cite{Yang2021,Wang2022}, which complicates the cross-comparison of methods. Such a state is in contrast to other ML fields. Computer vision has popular benchmarking datasets (such as ImageNet\footnote{\url{https://www.image-net.org}.}), and so does natural language processing (such as the Gutenberg project\footnote{\url{https://www.gutenberg.org}.}). These datasets are also included in ML libraries, such as PyTorch\footnote{\url{https://pytorch.org/vision/main/datasets.html}.} or TensorFlow\footnote{\url{https://www.tensorflow.org/datasets}.}, that offer the data pre-split into training and testing parts and loaded into suitable data structures. Such data preparation then accelerates the ML development and reduces the space for evaluation mistakes.

The arrival of the new multi-week CESNET datasets accelerated the need for supportive tools. The huge size of these datasets complicates the whole ML pipeline~\cite{horien2021hitchhiker}. It requires significant effort in data loading, optimization, and hardware resources, and is prone to errors. On the other hand, the increased effort allows complex time-aware evaluation on large datasets, bringing the important benefit of increased trust in the studied method and setting more accurate real-world performance expectations. DataZoo facilitates experiments with extensive data and contains options for time-aware and service-aware data splitting to simulate data drift events and novel service occurrences, enabling the use of novel datasets to their full potential.

% Time-aware evaluation measuring the performance degradation of a classification model requires a dataset spanning a longer period of time. The included CESNET datasets contain a large number of samples and span from two weeks to one year. However, the huge size of these datasets then complicates the whole ML pipeline~\cite{horien2021hitchhiker}. It requires significant effort in data loading, optimization, hardware resources, and is prone to errors. The increased effort brings an important benefit of increased trust in the studied method and sets more accurate real-world performance expectations. DataZoo contains time-aware and service-aware data splitting to simulate data drift events and novel service occurrences, bringing a more realistic evaluation to the network traffic classification field. 

The rest of the text is organized as follows. ~\secref{sec:datazoo} contains a description of the toolset, its configuration options, and details about time-aware and open-world evaluation. ~\secref{sec:conclusion} concludes the manuscript. 

\section{DataZoo toolset}
\label{sec:datazoo}
This section describes the core functions of the DataZoo toolset. Its main goal is to bring a standard API for handling datesets in the network traffic classification area. The toolset simplifies the downloading, configuring, and splitting of datasets, allowing faster ML development and better benchmarking of network traffic classifiers. 

\subsection{The current task in focus -- web service classification}
Traffic classification has a lot of sub-tasks, for example, mobile application classification~\cite{taylor_robust_2018, aceto_mobile_2019, rezaei_large-scale_2020}, malware detection~\cite{wang_malware_2017, anderson_deciphering_2018}, user behavior identification~\cite{li_app_2020}, VPN/tunnel detection~\cite{aceto_distiller_2021, draper-gil_characterization_2016}, or web service classification~\cite{Fauvel2023, shbair_multi-level_2016, akbari_look_2021}. The web service classification is the current focus of the toolset because there are multiple large public datasets available for this task. Support for other tasks, such as traffic categorization or mobile traffic classification, is planned in future releases.

% The initial release of DataZoo comes with three datasets targeting encrypted TLS and QUIC traffic. Detailed information about these datasets is written in~\tabref{tab:datasets}, and all of them share the following characteristics:

The initial release of DataZoo comes with three datasets targeting encrypted TLS and QUIC traffic. Detailed information about these datasets is provided in~\tabref{tab:datasets}. The process of data collection was the same for all three datasets and is described in~\cite{Luxemburk2023QUICDataset}. Datasets also share the following characteristics:

\begin{enumerate}
    \item Each sample represents a bidirectional network flow.
    \item Ground-truth labeling was done based on the Server Name Indication (SNI) domain that is sent during a connection handshake.
    \item Data features can be divided into two categories: packet sequences (sizes, inter-arrival times, directions) and flow statistics describing the entire connection (for example, the number of transmitted packets or the flow duration).
    \item Samples in the dataset are organized per date. This enables time-aware evaluation of traffic classifiers.
\end{enumerate}
 
\subsection{Common API}
DataZoo is implemented as a Python package. The package defines a common dataset interface and provides classes for individual datasets. Datasets are configured using a dedicated \texttt{DatasetConfig} class, which integrates  all the options related to train/validation/test splitting, application selection, and feature scaling. The configuration options are discussed in the next sections.

All the datasets are offered in four sizes: \texttt{XS} with 10M samples, \texttt{S} with 25M, \texttt{M} with 50M, \texttt{L} with 100M, and the original size. The smaller sizes give users an option to start experimenting at a smaller scale (also faster dataset download, lower disk space, etc.). The default is the \texttt{S} size containing 25 million samples.

\begin{figure}[h]
\begin{minted}[
    frame=lines,
    fontsize=\smaller,
    ]{python}
from cesnet_datazoo.datasets import CESNET_QUIC22
from cesnet_datazoo.config import (DatasetConfig,
                                   ValidationApproach)                             
dataset = CESNET_QUIC22(data_root="/data/CESNET-QUIC22/",
                        size="XS")
dataset_config = DatasetConfig(
    dataset=dataset,
    val_approach=ValidationApproach.SPLIT_FROM_TRAIN,
    train_period="W-2022-44",
    test_period="W-2022-45",
)
dataset.set_dataset_config_and_initialize(dataset_config)
train_dataframe = dataset.get_train_df()
val_dataframe = dataset.get_val_df()
test_dataframe = dataset.get_test_df()
\end{minted}
\captionof{listing}{Example code for obtaining train, validation, and test data.}
\label{code:example}
\end{figure}
\vspace{-0.3cm}
The selected dataset size is downloaded when initializing a dataset class (such as \texttt{CESNET\_QUIC22} from the example Listing~\ref{code:example}). Once the configuration is set with the \texttt{set\_dataset\_config\_and\textunderscore\allowbreak initialize} method, train, validation, and test data become available. The two options for accessing data are:

\begin{itemize}
    \item Iterable PyTorch \texttt{DataLoader}\footnote{\url{https://pytorch.org/docs/stable/data.html\#torch.utils.data.DataLoader}.} for batch processing.
    \item Pandas \texttt{DataFrame}\footnote{\url{https://pandas.pydata.org/docs/reference/api/pandas.DataFrame.html}.} for loading the entire train, validation, or test set into RAM at once.
\end{itemize}

Batch processing is needed for training some machine learning models, such as neural networks. Also, larger datasets cannot fit into RAM, and their processing needs to be chunked with the \texttt{DataLoader} interface.

\subsection{Time-aware evaluation}

\begin{table}[t]
    \centering
    \small
    % \linespread{0.9}\selectfont
    % \setlength\extrarowheight{2pt}
    \caption{Time-aware evaluation configuration options.}
    \label{tab:time-conf}
    \begin{tabular}{|P{2cm}|P{5.7cm}|}
    \hline
        \textbf{Name} & \textbf{Description}  \\ \hline
        Train dates & Which dates of the dataset should be used for creating the train set. \\ \hline
        Weighs of train dates & Used to provide a specific distribution of samples across train dates. \\ \hline
        Test dates & Which dates of the dataset should be used for creating the test set. \\ \hline
        Validation approach & Whether to use a fraction of train set as validation or define separate dates. \\ \hline
        Validation dates & Separate validation dates. \\ \hline
        Validation fraction & Fraction of validation samples when splitting from train. \\ \hline
    \end{tabular}
\end{table}

All the available datasets are built from traffic captured during longer periods of time, ranging from one week to one year. The samples in the datasets are organized per date, which gives users an option to choose what dates to use for training, validation, and testing. For creating a validation set, there are two options in the toolset. The first one, which is the default, is to split train data into train and validation. Scikit-learn \texttt{train\_test\_split}\footnote{\url{https://scikit-learn.org/stable/modules/generated/sklearn.model_selection.train_test_split.html}.} is used to create a random stratified validation set. The second option is to define a separate list of dates that should be used for creating a validation set. An overview of available dataset partitioning configuration options is in~\tabref{tab:time-conf}.

When a list of dates is specified for train, validation, or test sets, it means that all samples from the given dates will be included. Unless a user requires a specific smaller size of train, validation, or test sets, in which case a random subset is taken. The toolset provides an option for building a train set with a non-uniform distribution of samples across train dates. This is useful for evaluating different training strategies, such as putting more weight on samples from the most recent dates. An example use case is training a model on one month of data and wanting to use more samples from the most recent week because the most recent week contains the most current traffic characteristics that the model should capture and learn. However, including some older traffic up to one month old can also be advantageous.

\subsubsection{Importance of time-aware evaluation}
The production networks undergo constant change---congestions, rush hours, new applications, and protocol updates. To prepare traffic classification models for this environment, we have to evaluate them on test sets capturing these changes. This should allow researchers to measure how the models are susceptible---or robust---in the face of network traffic data drift. 

\begin{figure}[t]
    \centering
    \includegraphics[width=\linewidth]{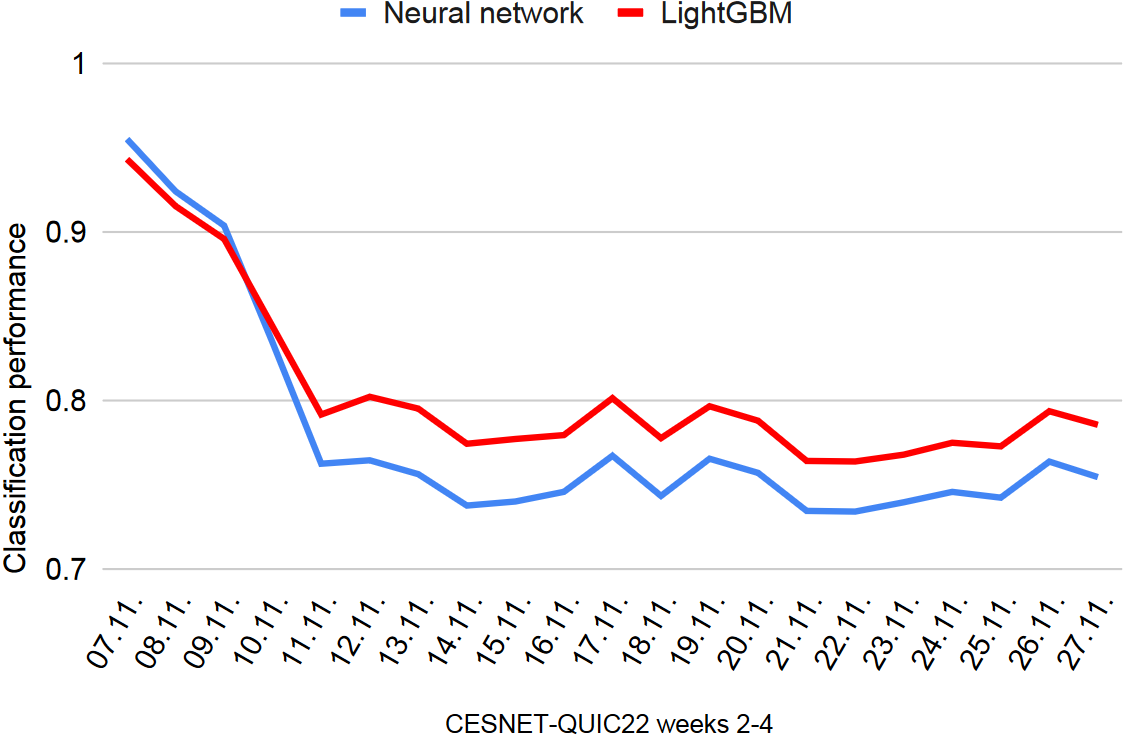}
    \caption{The classification performance of two ML models on CESNET-QUIC22 measured per date. We can see a noticeable performance drop due to data drift.}
    \label{fig:data-drift-example}
\end{figure}

There are only a handful of traffic classification studies performing time-aware evaluation. In a recent paper~\cite{Luxemburk2023QUIC}, the authors studied QUIC traffic classification and measured the performance changes of neural network and LightGBM models in time. Huge, up to 14\%, performance drops were discovered between the second and third weeks of the CESNET-QUIC22 dataset. The measured performance is visualized in~\figref{fig:data-drift-example}. Those drops were attributed to a change of the Google TLS server certificate. This is a good example of a data drift event that would not be discovered if the evaluation was done on shuffled data, without taking into account the time domain. Malekghaini et al.~\cite{Malekghaini2023} observed data drift on network traffic datasets years apart, which is not surprising given that the time gap between model training and evaluation is so long.

\subsection{Open-world setting}

\begin{table}[b]
    \centering
    \small
    \caption{Approaches for splitting application between \textit{known} and \textit{unknown}.}
    \label{tab:app-selection-options}
    \begin{tabular}{|P{2cm}|P{5.8cm}|}
    \hline
        \textbf{Application selection} & \textbf{Description}  \\ \hline
        All known & All applications are considered as \textit{known}. \\\hline
        Top-X & Use the top X most frequent (with the most samples) applications as \textit{known}, and the rest as \textit{unknown}. \\ \hline
        Explicit unknown & Manually specify a list of \textit{unknown} applications. Useful for the CESNET-QUIC22 dataset, which contains several background traffic classes. \\ \hline
        Fixed & Fixed \textit{known} and \textit{unknown} application selection for long-term multi-week evaluation. \\ \hline
    \end{tabular}
\end{table}

Another important aspect of the realistic evaluation of network traffic classifiers is to account for novel classes. Training datasets in the traffic classification domain cannot be complete in the sense of containing all possible classes. The Internet is a dynamic environment, and new apps and services are emerging each week. Traffic classification should thus be evaluated in changing environments with novel (unseen in the training set) classes.

Papers focusing on detecting novel, out-of-distribution (OOD) samples in network traffic are summarized in Table~\ref{tab:ood-papers}. In the current state, it is impossible to perform cross-comparison of methods for OOD detection in network traffic because each individual proposal uses a different dataset. Also, the OOD evaluation is even more prone to mistakes than the standard closed-world setting evaluation.

To address this issue and to enable better cross-comparison of OOD detection methods, the toolset provides multiple approaches to split classes between \textit{known} and \textit{unknown}. The main options are listed in~\tabref{tab:app-selection-options}. The default is to use all classes as \textit{known} (the standard closed-world setting).

\begin{table}[t]
    \small
    \centering
    \caption{Related work in network traffic OOD detection. The abbreviations stand for: \textit{G} -- Thinkback method based on gradients~\cite{yang_thinkback_2021}; \textit{E} -- Energy method~\cite{liu_energy-based_2021}; \textit{C} -- thresholding of the model confidence; TPR -- fraction of correctly identified \textit{unknown} samples; FPR -- fraction of \textit{known} samples incorrectly identified as \textit{unknown}; \#K, \#U -- number of \textit{known}, \textit{unknown} classes.}
    \label{tab:ood-papers}
    \begin{threeparttable}
    \setlength\extrarowheight{1pt}
    \begin{tabular}{|p{2.1cm}|l|l|p{27mm}|}
    \hline
        \textbf{Paper} & \textbf{M.} & \textbf{\#K + \#U} & \textbf{OOD performance} \\ \hline
        Yang et al.~\cite{Yang2021} & \textit{G} & 162 + 500  & 78\% TPR at 1\% FPR\ \\ \hline
        Luxemburk\;et al.~\cite{Luxemburk2023} & \textit{E}, \textit{G} & 100 + 91  & \textit{E}: 88\% TPR at 5\% FPR \newline \textit{G}: 92\% TPR at 5\% FPR \\ \hline
        Hu et al.~\cite{hu2022} & FAE-G & 16 + 4 & 58\% TPR at 3.5\% FPR \\ \hline
        Dahanayaka\;et al.~\cite{DAHANAYAKA2023109991} & k-LND & 39 + 59\tnote{a} & 76\% TPR at 4\% FPR\tnote{b}\\ \hline
        Aceto et al.~\cite{aceto_mobile_2019} & \textit{C} & only \textit{known} & not measured\tnote{c} \\ \hline
        Taylor et al.~\cite{taylor_robust_2018} & \textit{C} & only \textit{known} & not measured \\ \hline

    \end{tabular}
    \begin{tablenotes}
      \item[a]{\footnotesize The evaluation was performed on five datasets; results of an IoT voice fingerprinting task are reported.}
      \item[b]{\footnotesize FPR was not directly provided in the paper. This is its upper bound computed from available metrics.}
      \item[c]{\footnotesize In evaluations without separate \textit{unknown} classes, the OOD performance cannot be properly measured. What can be reported is the fraction of rejected \textit{known} samples.}
    \end{tablenotes}
  \end{threeparttable}
\end{table}

\subsection{Feature scaling}
Scaling of input data is recommended for some models, such as neural networks, but not required for others, such as XGBoost. Thus, the toolset provides optional and configurable feature scaling. During the initialization of the train set, its fraction is used for fitting three separate scalers---for packet sizes, inter-packet times, and flow statistics. These three Scikit-learn\footnote{\url{https://scikit-learn.org/stable/modules/classes.html\#module-sklearn.preprocessing}.} scalers are available: \texttt{StandardScaler}, \texttt{RobustScaler}, and \texttt{MinMaxScaler}. There are also options to perform min and max clipping (for packet sizes and inter-packet times) or quantile clipping (for flow statistics) before data scaling to reduce the influence of outliers. The fitted scalers are cached and reused for future experiments with compatible dataset configuration.

\begin{table}[t]
    \centering
    \small
    \caption{Feature scaling options.}
    \label{tab:data-scaling-options}
    \begin{tabular}{|P{2cm}|P{5.8cm}|}
    \hline
        \textbf{Name} & \textbf{Description}  \\ \hline
        Fit scalers samples & Fraction of the train set used for fitting scalers \\ \hline
        Packet sizes scaler & Standard, robust, min-max, or no scaling of packet sizes \\ \hline
        Packet sizes max clip & Before scaling, packet sizes larger than this value are replaced with this value \\ \hline
        IPT scaler & Standard, robust, min-max, or no scaling of inter-packet times (IPT) \\ \hline
        IPT max clip & Before scaling, IPT larger than this value are replaced with this value \\ \hline
        IPT min clip & Before scaling, IPT smaller than this value are replaced with this values\\ \hline
        Flow statistics scaler & Standard, robust, min-max, or no scaling of flow statistics \\ \hline
        Flow statistics quantile clip $q$ & Before scaling, flow statistics greater than their $q$-quantile are clipped \\ \hline
    \end{tabular}
\end{table}
 
\section{Conclusion}
\label{sec:conclusion}

The lack of representative benchmarking datasets slows progress in the network traffic classification field~\cite{Wang2022,Yang2021}. In other ML areas, standard repositories and dataset APIs exist and are used to accelerate research. Therefore, we created the DataZoo toolset to bring a standard data-management tool into network traffic classification. DataZoo contains three extensive datasets targeting web service classification and supportive functions automatizing data-handling procedures. Moreover, the toolset simplifies evaluations that take into account the dynamic nature of computer networks. The provided time-aware and service-aware data splitting allow researchers to simulate data drifts and novel service appearances, which in turn enables setting more accurate real-world performance expectations.

The DataZoo toolset is documented\footnote{\url{https://cesnet.github.io/cesnet-datazoo/}.} and can be installed from PyPI\footnote{\url{https://pypi.org/project/cesnet-datazoo/}.} or GitHub\footnote{\url{https://github.com/CESNET/cesnet-datazoo}.}. We plan to continue in the development and add support for additional popular datasets, such as AppClassNet~\cite{Wang2022} or ISCXVPN2016~\cite{draper-gil_characterization_2016}. Moreover, we encourage researchers to cooperate and share tools and datasets in order to accelerate progress and improve the overall state of machine learning evaluations across the network traffic classification field.

\section*{Acknowledgements}
This work was supported by the Ministry of the Interior of the Czech Republic, grant No. VJ02010024: \textit{``Flow-Based Encrypted Traffic Analysis,''} and also by the Grant Agency of the Czech Technical University in Prague, which is funded by the Ministry of Education, Youth and Sports of Czech Republic, grant No. SGS23/207/OHK3/3T/18.

\bibliographystyle{ACM-Reference-Format}
\bibliography{references.bib}

\end{document}